\documentclass{article}

\PassOptionsToPackage{numbers, compress}{natbib}

\usepackage[preprint]{neurips_2026.arxiv}

\usepackage[utf8]{inputenc}
\usepackage[T1]{fontenc}
\usepackage{hyperref}
\usepackage{url}
\usepackage{booktabs}
\usepackage{multirow}
\usepackage{subcaption}
\usepackage{amsfonts}
\usepackage{amsmath}
\usepackage{amssymb}
\usepackage{nicefrac}
\usepackage{microtype}
\usepackage{xcolor}
\usepackage{graphicx}
\usepackage{tikz}
\usetikzlibrary{positioning,arrows.meta,shapes.geometric,fit,calc}
\usepackage{algorithm}
\usepackage{algpseudocode}
\usepackage[capitalize,noabbrev]{cleveref}

\graphicspath{{figures/}}

\newcommand{\microtask}{micro-task}
\newcommand{\Microtask}{Micro-task}
\newcommand{\microtasks}{micro-tasks}

\newcommand{\rmtl}{RMTL}

\title{RMTL: Reinforced Micro-task Learning for Long-Horizon Manipulation with VLM Rewards}

\author{%
  An{\i}l Can Ate\c{s}\thanks{Corresponding author.} \\
  Istanbul Technical University \\
  \texttt{atesa19@itu.edu.tr}
  \And
  Orhan Kahraman \\
  Istanbul Technical University \\
  \texttt{kahramano20@itu.edu.tr}
  \And
  Cihan Topal \\
  Istanbul Technical University \\
  \texttt{cihantopal@itu.edu.tr}
}

\begin{document}

\maketitle

\begin{figure}[h!]
  \centering
  \begin{tikzpicture}[
    every node/.style={font=\small},
    axisline/.style={->, line width=0.45pt, gray!75},
    arr/.style={->, line width=0.45pt, gray!60},
    pbox/.style={draw, rounded corners=1.2pt, minimum height=4mm,
                 font=\scriptsize, inner sep=2pt, align=center},
  ]
    \begin{scope}
      \node[anchor=south west, font=\bfseries\small] at (0, 1.95)
        {(a) Single global prompt};
      \draw[axisline] (0, 0) -- (4.4, 0);
      \draw[axisline] (0, 0) -- (0, 1.75);
      \node[font=\scriptsize, anchor=north east] at (4.4, -0.02)
        {env step};
      \node[font=\scriptsize, rotate=90, anchor=south] at (-0.20, 0.85)
        {VLM reward};
      \draw[blue!65, line width=1.0pt, smooth, tension=0.55]
        plot coordinates {
          (0.0, 0.20) (0.5, 0.18) (1.0, 0.22) (1.5, 0.24)
          (2.0, 0.20) (2.5, 0.24) (3.0, 0.32) (3.5, 0.62) (4.0, 1.05)
        };
      \node[font=\scriptsize, gray!75, align=center]
        at (1.7, 1.20) {near-flat early\\ \emph{(no useful gradient)}};
      \draw[arr] (1.7, 0.95) -- (1.55, 0.32);
      \node[pbox, fill=blue!7, draw=blue!50] at (2.1, -0.65)
        {"a robot gripper lifting a red cube"};
    \end{scope}
    \begin{scope}[xshift=5.6cm]
      \node[anchor=south west, font=\bfseries\small] at (0, 1.95)
        {(b) \rmtl{} (ours): per-\microtask{} prompts};
      \draw[axisline] (0, 0) -- (4.4, 0);
      \draw[axisline] (0, 0) -- (0, 1.75);
      \node[font=\scriptsize, anchor=north east] at (4.4, -0.02)
        {env step};
      \draw[gray!35, dashed, line width=0.4pt] (1.45, 0) -- (1.45, 1.75);
      \draw[gray!35, dashed, line width=0.4pt] (2.85, 0) -- (2.85, 1.75);
      \draw[blue!75, line width=1.15pt, smooth, tension=0.5]
        plot coordinates {(0.0, 0.10) (0.4, 0.30) (0.8, 0.55)
                          (1.2, 0.78) (1.45, 0.86)};
      \draw[orange!85!black, line width=1.15pt, smooth, tension=0.5]
        plot coordinates {(1.45, 0.86) (1.85, 0.94) (2.30, 1.04)
                          (2.65, 1.12) (2.85, 1.16)};
      \draw[red!80!black, line width=1.15pt, smooth, tension=0.5]
        plot coordinates {(2.85, 1.16) (3.20, 1.27) (3.60, 1.42)
                          (4.05, 1.57)};
      \node[pbox, fill=blue!8,    draw=blue!55]        at (0.72, -0.55)
        {\textit{approach}};
      \node[pbox, fill=orange!14, draw=orange!75!black] at (2.15, -0.55)
        {\textit{align}};
      \node[pbox, fill=red!11,    draw=red!70!black]   at (3.45, -0.55)
        {\textit{grasp}};
      \node[font=\scriptsize, gray!75, align=center]
        at (2.15, 1.55) {piecewise-monotone signal};
    \end{scope}
  \end{tikzpicture}
  \caption{\textbf{\rmtl{} at a glance.}
    A single task-level VLM prompt can produce a nearly flat reward over early parts of a long-horizon manipulation trajectory. RMTL instead decomposes the task into stage-specific language prompts and evaluates the agent using only the currently active micro-task prompt. This produces more progress-aligned reward variation within each stage.}
  \label{fig:teaser}
\end{figure}
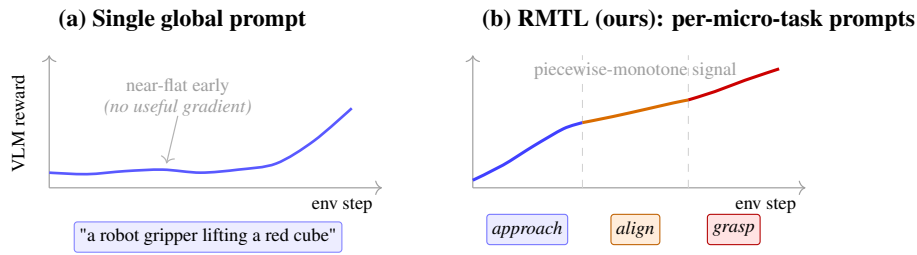
\begin{abstract}
Reinforcement learning (RL) for robotic manipulation often requires manually designing a dense reward function, which is difficult to tune and often fragile, or learning a reward from human demonstrations or preferences, which can be expensive. 
A recent line of work uses pretrained vision-language models (VLMs) as zero-shot reward models, replacing these costs with a single text prompt. However, we argue
that a single global prompt is too coarse for long-horizon manipulation tasks with randomized initial conditions. 
The single-prompt VLM reward is near-flat for much of the trajectory, making early progress hard for the agent to detect. 
We propose \textbf{R}einforced \textbf{M}icro-\textbf{T}ask \textbf{L}earning (\rmtl{}), an approach that decomposes a manipulation task into a small set of language-described \microtasks{} and trains the agent to switch between them. 
At each step, the agent receives a multi-view VLM reward computed using the prompt of the currently active \microtask{} and averaged across multiple camera views to reduce the effect of view-specific occlusions.
A reverse curriculum gradually exposes the agent to harder initial conditions, while a PPO worker is first trained with a fixed distance-based rule that selects the active \microtask{}. 
We then replace this rule with a learned hierarchical manager, turning rule-based phase selection into a fully learned hierarchical policy. 
We instantiate \rmtl{} on the Fetch manipulation environment using three short stage-specific prompts and without additional prompt tuning. 
Experiments show that \rmtl{} provides more informative reward signals than single-prompt VLM rewards, enabling faster learning.
These results suggest that decomposing VLM rewards into \microtask{}-specific language prompts can substantially improve the scalability of language-guided reinforcement learning for robotic manipulation.
\end{abstract}



\section{Introduction}
\label{sec:introduction}

Designing a dense reward function for a long-horizon manipulation task is notoriously hard. 
Hand-engineered shaping signals are brittle and rarely transfer across embodiments or tasks; reward learning from human demonstrations or preferences is expensive and difficult to scale. 
A recent line of work circumvents both options by using a frozen pretrained vision-language model (VLM) as a zero-shot reward model~\citep{rocamonde2024vlmrm,sontakke2023roboclip,ma2024eureka}. 
The agent receives, for each rendered frame, a similarity score between the frame's image embedding and a short natural-language description of the desired outcome.

\paragraph{The single-prompt failure mode.} While appealing, a single global prompt that describes the \emph{final} goal state is poorly suited to long-horizon control with randomized initial conditions. 
Empirically the language reward is near-flat for most of the trajectory: the gripper spends many steps far from any frame configuration that the prompt describes, so the embedding similarity changes only marginally as it moves.
At the same time, any one camera is hostage to gripper- and cube-induced occlusion: a perspective that conveniently shows the cube one moment can be entirely blocked by the gripper the next. Both effects waste signal.

\paragraph{Contributions.}
We propose \textbf{Reinforced \Microtask{} Learning (\rmtl{})}, a pipeline that decomposes a manipulation task into a small set of language-described \microtasks{} and trains an agent to switch between them, with VLM rewards aggregated over multiple camera views and a reverse curriculum on initial conditions (\Cref{fig:teaser}). Concretely, we contribute the following:
\vspace{-2mm}
\begin{enumerate}\itemsep1pt
  \item We show that decomposing the VLM reward into \textbf{three short, stage-specific language prompts} (\emph{approach}, \emph{align}, \emph{grasp}) turns a near-flat single-prompt signal into a piecewise-monotone signal that PPO can climb, on the same
    environment and same VLM (\S\ref{sec:reward-formula},\,\ref{sec:expt-results};
    \Cref{fig:subtask_vs_onetask}).
  \item We show that a small \textbf{learned hierarchical manager}, BC-warmstarted from a rule-based selector and refined by REINFORCE over a frozen PPO worker, matches or exceeds the rule while remaining stable (\S\ref{sec:manager},\,\ref{sec:expt-results}; \Cref{fig:behavioural_4way}).
  \item We show that \textbf{multi-view aggregation} of per-camera VLM scores reduces per-view occlusion noise and yields a smoother training signal than any single view, with no gradient-shape penalty (\S\ref{sec:why-multiview}; \Cref{fig:multiview_snap,fig:multiview_story}).
  \item We show that a \textbf{reverse curriculum} is the necessary glue: without it the language reward is too uninformative at the random-init endpoint to bootstrap PPO (\S\ref{sec:rc},\,\ref{sec:expt-results}; \Cref{fig:rc_progression}).
\end{enumerate}
\vspace{-2mm}

The remainder of the paper introduces related work (\S\ref{sec:related}), develops the method (\S\ref{sec:method}), presents experiments (\S\ref{sec:experiments}), and discusses limitations and future work (\S\ref{sec:discussion},\,\ref{sec:future}).

\section{Related work}
\label{sec:related}

\paragraph{RL for robotic manipulation.}
Reinforcement learning in simulation has produced standard benchmarks for contact-rich manipulation. 
The Fetch family of mobile-manipulator tasks introduced by~\citet{plappert2018gymrobotics} highlights the recurring difficulty of reward formulation: in pick-and-place, agents typically rely on dense hand-designed geometric rewards or on sparse binary success signals. 
Sparse rewards demand excessive exploration, while dense geometric shaping is sensitive to local minima and assumes access to privileged simulator state. 
Our work removes this privileged state from the reward path and instead derives a dense reward purely from rendered visual observations.

\paragraph{Pixel-based and visual RL.}
Moving from state to pixel observations introduces a sample-efficiency gap that has been partially closed by representation learning. RAD \citep{laskin2020rad} shows that diverse image augmentations alone let pixel-based RL match state-based methods, and CURL \citep{srinivas2020curl} achieves the same with contrastive self-supervision. 
These methods improve the encoder of the policy but still rely on environment-provided task rewards. 
We are orthogonal: the environment provides no intrinsic task reward and the visual observation \emph{is} the reward, computed by an external frozen VLM.

\paragraph{VLMs as reward models.}
A growing body of work uses pretrained vision-language models to score RL trajectories.~\citet{rocamonde2024vlmrm} introduce VLM-RMs, a family of zero-shot rewards that contrast a target prompt against a baseline using CLIP~\citep{radford2021clip}; we build directly on this projected scorer and extend it to per-\microtask{} prompts and multi-view aggregation. 
\citet{sontakke2023roboclip} use a video-text encoder for one-shot imitation.
\citet{ma2024eureka} treat the VLM as a code generator for hand-shaped reward functions. 
Other lines pretrain representations whose embeddings double as rewards: VIP \citep{ma2022vip} derives a value-implicit embedding from human videos, LIV \citep{ma2023liv} adds language-image rewards from action-free video, and~\citet{fan2022minedojo} train an internet-scale contrastive reward model for language-conditioned tasks in Minecraft. 
Across all of these the reward is conditioned on a \emph{single, global} language prompt describing the final goal, which is the regime in which we identify the near-flat-reward failure mode and which our \microtask{} decomposition is designed to address.

\paragraph{Hierarchical RL and task decomposition.}
Hierarchical methods that decouple high-level skill selection from low-level control are a long-standing line of research; the options framework of~\citet{sutton1999options} formalises the multi-level control structure, and option-critic~\citep{bacon2017optioncritic} provides an end-to-end gradient. For continuous control, HIRO \citep{nachum2018hiro} successfully applies a goal-conditioned two-level hierarchy. A long-standing open question in this line is how to discover sub-goals automatically. Our \rmtl{} manager grounds sub-goals directly in language: a small (\(\approx 2\)k-parameter), discrete-action manager selects a \microtask{} prompt and is trained with REINFORCE on segment returns. Stage 1 supplies the manager with a rule-based teacher; Stage 2 replaces that rule with the learned manager itself, turning heuristic phase selection into a fully learned policy.

\paragraph{Language-conditioned policies vs.\ language as reward.}
Beyond reward generation, language has been used directly to condition control policies and ground robotic affordances. RT-1 \citep{brohan2022rt1} tokenises language instructions and camera images and outputs motor commands; SayCan~\citep{ahn2022saycan} grounds the semantic knowledge of LLMs in pretrained robotic skills so that high-level planners can sequence them. These methods rely heavily on demonstration datasets and imitation learning. Our setting
is purely RL: language is used \emph{only} as a supervisory signal for the reward, and policies are trained from scratch in simulation without expert demonstrations.

\paragraph{Curriculum and reverse curriculum learning.}
Reverse curricula~\citep{florensa2017reversecurriculum} schedule the distribution of initial conditions from easy to hard. We adopt a scalar form: a single difficulty knob \(d \in [0, 1]\) interpolates the gripper between an "above-cube" trivial start and a fully randomized workspace init, and is gated by per-stage success thresholds.

\paragraph{Imitation-warmstarted policies.}
Behaviour cloning followed by RL is a common recipe~\citep{ross2011dagger}; we use it for the manager (not the worker), with rule-based teacher labels available essentially for
free.

\section{Method}
\label{sec:method}

\begin{figure}[!t]
    \centering
    \includegraphics[width=0.93\linewidth]{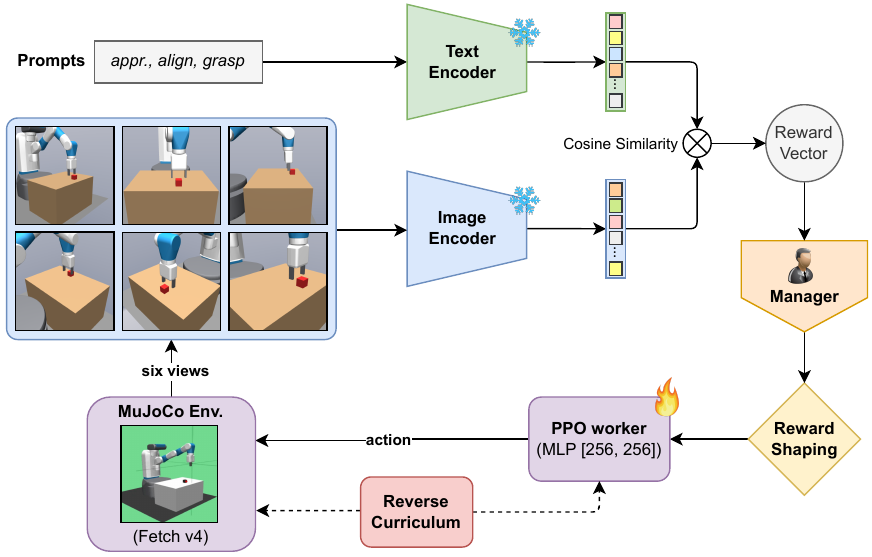}
    \caption{\rmtl{} data flow. Solid arrows are data; dashed arrows are selection/control. The phase-aware scorer chooses which target embedding to use based on the active \microtask{}, supplied either by a distance-based rule (Stage 1) or by a learned manager (Stage 2).}
  \label{fig:pipeline}
  \vspace{-2mm}
\end{figure}

We organize the method around the data flow shown in \Cref{fig:pipeline}: at each step the environment renders a small bank of camera views, a frozen VLM scores each view against the prompt of the currently active \microtask{}, the per-view scores are mean-aggregated into a scalar reward, and a PPO worker uses that reward to update its policy. The active \microtask{} is selected either by a fixed distance-based rule (Stage 1) or by a learned manager (Stage 2). A reverse curriculum schedules the distribution of initial gripper configurations.

\subsection{Preliminaries and problem formulation}
\label{sec:problem-setup}

We formulate robotic manipulation as an episodic finite-horizon reinforcement-learning problem. 
The interaction between the robot and the environment is modelled as a Markov decision process \(\mathcal{M} = (\mathcal{S}, \mathcal{A}, P, \rho_0, H, \gamma)\), with state space \(\mathcal{S}\), continuous action space \(\mathcal{A}\), transition dynamics \(P(s_{t+1}\mid s_t, a_t)\), initial-state distribution \(\rho_0\), horizon \(H\), and discount factor \(\gamma\). At each time step the agent observes \(o_t = O(s_t)\), samples \(a_t \sim \pi_\theta(\cdot \mid o_t)\) and receives a scalar reward \(r_t\). The policy is trained to maximize the expected discounted return:
\begin{equation}
  J(\theta)
  = \mathbb{E}_{s_0 \sim \rho_0,\, a_t \sim \pi_\theta}
    \left[\sum_{t=0}^{H-1} \gamma^t r_t\right].
  \label{eq:rl-objective}
\end{equation}
In manipulation, specifying \(r_t\) is the main bottleneck. Hand-designed dense rewards are brittle and require privileged simulator states, while sparse success rewards make exploration unfeasible in long-horizon tasks where terminal states are rarely reached by an untrained policy.

We instead consider rewards derived from a \emph{frozen} pretrained vision-language model. 
Let \(I_t = \mathcal{R}(s_t)\) denote a rendered image of the environment state (or a tuple of such images, one per camera view), and let \(F_\psi(I_t, p)\) be the scalar compatibility score assigned by a VLM with fixed parameters \(\psi\) to image \(I_t\) and text prompt \(p\).
The policy parameters \(\theta\) are optimized; \(\psi\) is never updated. 
In our implementation \(F_\psi\) is instantiated as a CLIP-style projected-embedding reward (\S\ref{sec:reward-formula}), but the rest of the method only assumes access to a scalar image-text alignment score.

A direct approach is to use a single global prompt \(p_G\) describing the final task outcome,
\begin{equation}
  r_t^G = F_\psi(I_t,\, p_G).
  \label{eq:single-global-reward}
\end{equation}
For example, \(p_G\) may describe a robot gripper lifting a cube. 
This formulation is simple, but it assumes that similarity to the final goal prompt provides a meaningful dense measure of progress \emph{throughout} the trajectory. In long-horizon manipulation that assumption is violated: the robot must pass through qualitatively different stages (reaching, aligning, grasping, lifting), and early states---necessary for eventual success---may have weak semantic similarity to the final goal prompt. 
The single-prompt reward then becomes nearly flat over large portions of the trajectory, leading to poor credit assignment and weak exploration guidance, especially under randomized initial conditions where most early interactions otherwise carry no learning signal.

The problem we address is therefore: given a manipulation MDP, a frozen VLM, and natural-language task descriptions, define a reward that is visually grounded, language-conditioned, and \emph{sensitive to intermediate task progress} across the full trajectory.

\subsection{Aggregate multi-view semantic scoring}
\label{sec:why-multiview}

Manipulation scenes are often partially occluded or ambiguous from a single viewpoint: the gripper may hide the cube, a close-up may omit global pose context, and a wide view may sacrifice fine contact detail.
We therefore compute the VLM score independently for each of $V{=}6$ fixed viewpoints and take the arithmetic mean as the scalar reward fed to PPO. This \emph{aggregate} score is a simple way to trade off complementary visual evidence without fusing pixels beforehand.

\begin{figure}[t!]
  \centering
  \includegraphics[width=0.85\linewidth]{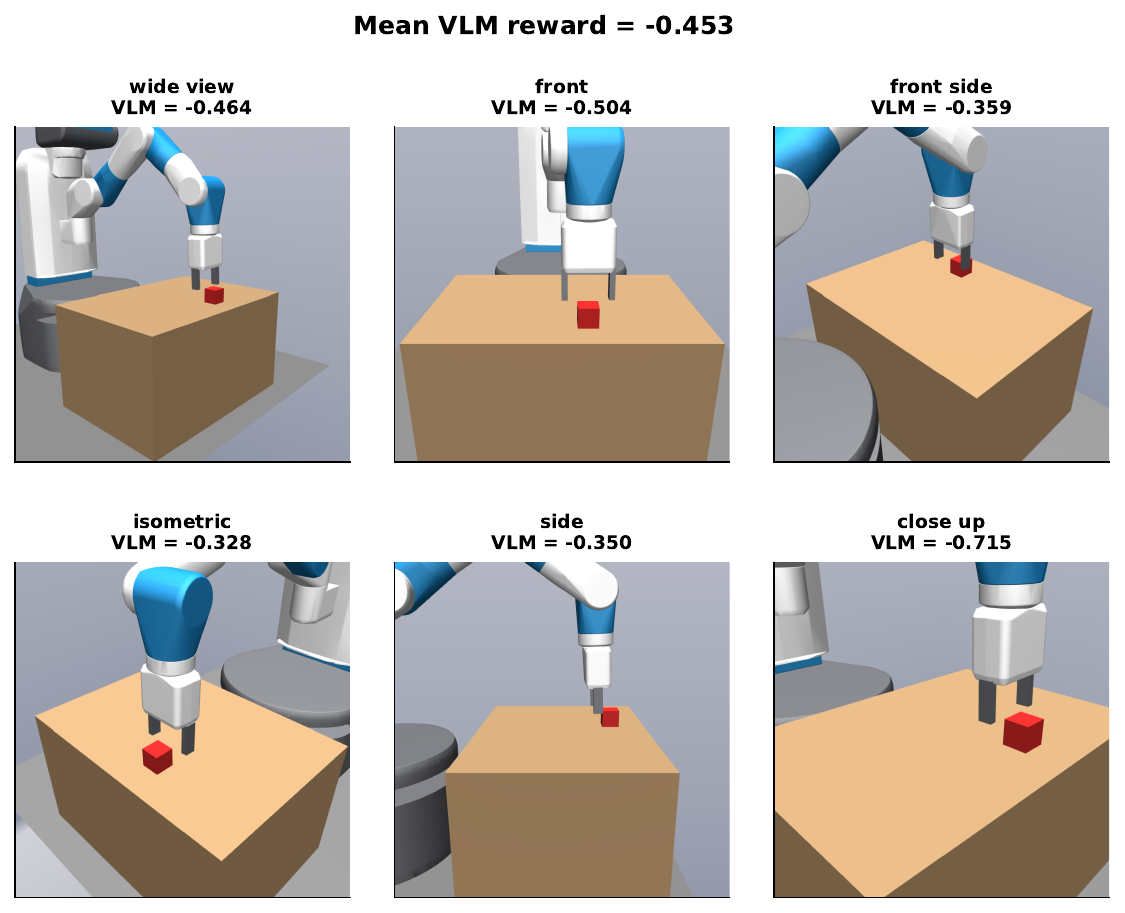}
  \caption{\textbf{Per-view vs.\ aggregate reward on one frame.}
    Six viewpoints of the same scene yield different VLM scores; their mean is the training signal.}
  \label{fig:multiview_snap}
  \vspace{-2mm}
\end{figure}

\Cref{fig:multiview_snap} makes the disagreement concrete on one frame: all six cameras see the same instantaneous state, yet their active-\microtask{} scores differ substantially. The mean smooths these per-view fluctuations while preserving the coarse semantic trend toward the prompt. Over full episodes, two independent rollouts in \Cref{fig:multiview_story} show the same pattern: the multi-view mean tracks a smoother trajectory than any single camera.

\subsection{Phased VLM reward}
\label{sec:reward-formula}

We score images using the projection-style scorer of \citet{rocamonde2024vlmrm} adapted to per-\microtask{} targets. 
Let $E$ denote the (fixed) image encoder and $T$ the text encoder of a VLM. 
We encode two groups of prompts once per training run:
\begin{itemize}
  \item a baseline prompt $b$ (a frame with no task progress, e.g.\ "a
    robot gripper and a red cube on a table"),
  \item a set of $K$ \microtask{} target prompts
    $\{\tau_i\}_{i=1}^K$ (one per \microtask{}, e.g.\ "a robot gripper
    pinching a red cube between two fingers").
\end{itemize}
Their $\ell_2$-normalised text embeddings are denoted $e_b = T(b)$ and $e_{\tau_i} = T(\tau_i)$. For \microtask{} $i$, the direction $d_i = e_{\tau_i} - e_b$ identifies the semantic axis from "no progress" to "\microtask{} $i$ satisfied"; the projection
\begin{align}
  P_\alpha(d_i) &= \alpha \,
    \frac{d_i d_i^{\top}}{\lVert d_i \rVert_2^2}
    + (1 - \alpha) \, I,
  \label{eq:projection}
\end{align}
trades off between rewarding only motion along that axis ($\alpha \to 1$) and unrestricted cosine similarity to $e_{\tau_i}$ ($\alpha \to 0$). 
For each camera view $v$ we compute a per-view image embedding $x_v = E(I_v)$
and score
\begin{align}
  r_v^{(i)} &= 1 - \tfrac{1}{2}\,
    \bigl\lVert (x_v - e_{\tau_i})\, P_\alpha(d_i) \bigr\rVert_2^2,
  \label{eq:per-view-reward}
\end{align}
and aggregate over views by the mean
\begin{align}
  r_t^{(i)} &= \frac{1}{V} \sum_{v=1}^{V} r_v^{(i)}.
  \label{eq:multiview-mean}
\end{align}
We use PE-Core-bigG-14-448~\citep{tschannen2025pecore} as the VLM, with $\alpha = 0.9$; the analysis that motivates this choice is given in \Cref{app:alpha-sweep}. The scorer in~\citet{rocamonde2024vlmrm} also admits an optional \emph{negative}-prompt term that subtracts cosine similarity to a distractor prompt; we did not find it necessary in the final configuration.

\subsection{Reverse curriculum}
\label{sec:rc}

Training directly under the hardest randomised resets is brittle: the gripper can start far enough from meaningful scene structure that the VLM provides almost no similarity gradient, so exploration never connects to success. 
We therefore adopt a \emph{reverse curriculum} that gradually widens the distribution of initial gripper (and finger) poses from an ``easy'' region (near the object, structured approach) toward the full randomised workspace draw used at convergence. 
The schedule is organised into five difficulty \emph{levels}; each level exposes strictly more challenging initial conditions than the previous one.
Progression is governed by online success statistics so that the agent is only exposed to the next level once it reliably solves the current one.
Formally, this is the same reverse-curriculum principle as \citet{florensa2017reversecurriculum}, specialised to randomised grippers in manipulation. 
Full numeric defaults (stage ranges, gating statistics, minimum dwell time) are listed in \Cref{app:impl}.

\subsection{Stage 1: PPO worker with rule-based \microtask{} switching}
\label{sec:rule-worker}

We instantiate \rmtl{} with $K = 3$ \microtasks{} aligned with the phases of pick-and-place: \emph{approach}, \emph{align}, and \emph{grasp}, each associated with one short target prompt (full prompt text in \Cref{app:impl}). 
At every step a deterministic rule selects the active \microtask{}; the phase-aware scorer indexes the corresponding $\tau_{i_t}$ in Eq.~\eqref{eq:per-view-reward}, and the worker receives the VLM reward of the currently active \microtask{} only --- never a sum or average across the three. 
The worker is trained with PPO~\citep{raffin2021sb3,schulman2017ppo}; full hyperparameters are listed in \Cref{tab:hparams_stage1}.

\subsection{Stage 2: learned hierarchical manager}
\label{sec:manager}

Stage 2 replaces the rule-based selector of \S\ref{sec:rule-worker} with a learned manager $\pi_M(\phi \mid s)$: a small MLP with a $K$-way softmax over the same three \microtasks{}, shared across all parallel environments. 
At every $H = 3$ environment steps the manager takes one decision $\phi \in \{0, 1, 2\}$, which is broadcast to the worker as a phase-onehot input and to the reward pipeline as the index for $\tau_{\phi}$ in Eq.~\eqref{eq:per-view-reward}. 
The manager is BC-warmstarted from rule-based labels and then refined by REINFORCE on segment returns; the worker is loaded from a Stage-1 PPO checkpoint and its actor is frozen for the first $50{,}000$ environment steps to stabilise the manager's REINFORCE target during the most non-stationary phase of training, then thawed at a reduced learning rate. 
The full procedure (BC warmstart, REINFORCE objective, KL anchor, $\varepsilon$-greedy exploration, freeze schedule, and the Stage-2 data-flow diagram) is given in \Cref{app:impl}, with hyperparameters in \Cref{tab:hparams_stage2} and the end-to-end algorithm in \Cref{alg:hrl-stage2}. 
Joint manager-and-worker training from scratch is left to future work (\S\ref{sec:future}).

\section{Experiments}
\label{sec:experiments}

\subsection{Setup}
\label{sec:expt-setup}

\paragraph{Environment.} We instantiate \rmtl{} on the
\texttt{FetchPickAndPlace-v4} environment from
Gymnasium-Robotics~\citep{plappert2018gymrobotics}, a 7-DoF Fetch arm
tasked with picking up a cube and holding it stably with the gripper
closed; physics is simulated by MuJoCo~\citep{todorov2012mujoco}. The
environment exposes a 25-dimensional state observation $s_t$
comprising gripper pose, cube pose, finger opening, and goal location.
Actions are 4-dimensional: gripper $\Delta xyz$ plus finger opening.
Episode success is detected geometrically once the gripper has stably
grasped and lifted the cube above the table (full criterion in
\Cref{app:impl}). Episodes are at most 40 steps long. Object, goal,
and gripper positions are randomized within workspace bounds at every
reset; the gripper's initial spatial spread is then modulated by the
reverse curriculum (\S\ref{sec:rc}).

\begin{figure}[!t]
  \centering
  \includegraphics[width=0.7\linewidth]{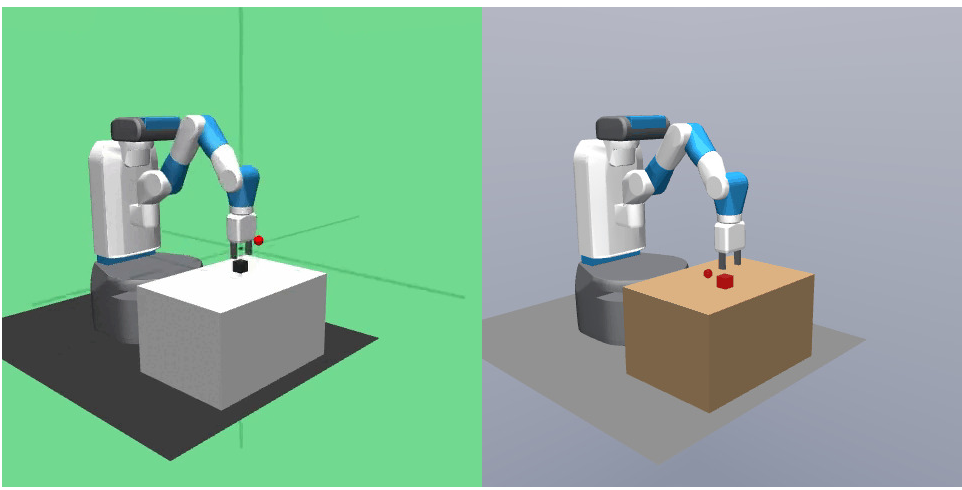}
  \caption{\textbf{Visual-domain modifications applied to the
    \texttt{FetchPickAndPlace-v4} renderer.} The manipulated cube is
    rendered in saturated red, the table is given a wooden material,
    and the background is set to neutral grey, with lighting tuned for
    contrast.}
  \label{fig:domain_realization}
  \vspace{-4mm}
\end{figure}

\paragraph{Visual domain.} Pretrained VLMs are trained primarily on
natural images and read raw MuJoCo renderings poorly: low-texture
surfaces, ambiguous object colors and visual artefacts make
image-text alignment unreliable. To narrow this domain gap we apply a
high-fidelity rendering setup: the manipulated cube is rendered in a
saturated red, the table is given a wooden appearance, the floor and
background are set to neutral grey tones, and lighting and material
parameters are tuned for contrast and object separation
(\Cref{fig:domain_realization}). These changes do not alter the task
dynamics or success condition; they only make the rendered scene more
semantically legible to the frozen VLM. The rendering setup is
identical across all methods we compare.

\paragraph{Compute.}
We train every variant on a single NVIDIA A100 GPU with 16 parallel
CPU environments per learner. Stage-1 runs typically converge near
600k environment steps; Stage-2 manager runs are trained for an
additional 200k environment steps starting from a loaded Stage-1
checkpoint.

The complete training setup for the Stage-1 PPO worker is summarised in
\Cref{tab:hparams_stage1}; Stage-2 hyperparameters specific to the
learned hierarchical manager are listed in \Cref{tab:hparams_stage2}.
Hyperparameters not repeated in \Cref{tab:hparams_stage2} (environment,
multi-view rendering, VLM scorer, reward shaping, RC, PPO worker) are
inherited unchanged from \Cref{tab:hparams_stage1}.

\subsection{Results}
\label{sec:expt-results}

We report all curves as eval success rate against environment steps.
\vspace{-2mm}

\begin{figure}[!b]
    \vspace{-2mm}
  \centering
  \includegraphics[width=0.5\linewidth]{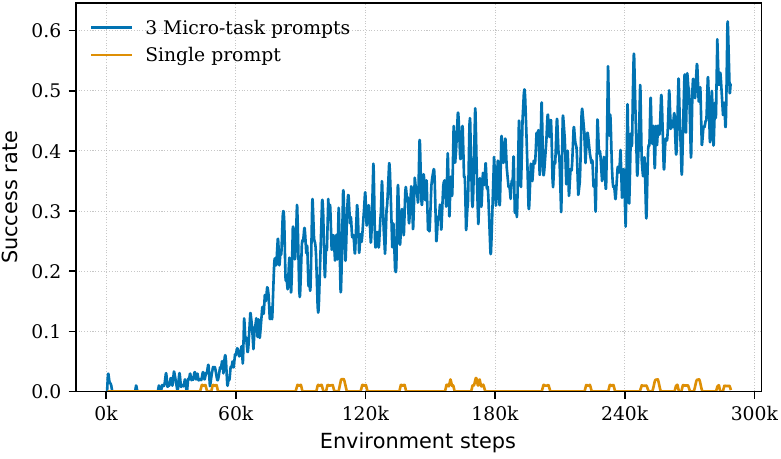}
  \caption{\textbf{Micro-task prompts vs.\ single-prompt ablation.} Success rate for full \rmtl{},
    multi-task prompts + multi-view, and single global prompt +
    multi view, with all other RL settings held fixed.}
  \label{fig:subtask_vs_onetask}
\end{figure}

\paragraph{Micro-task vs.\ single-prompt baseline.}
\Cref{fig:subtask_vs_onetask} compares \rmtl{} (three per-\microtask{}
prompts + multi-view) against two strict counterparts that share the
exact same RL hyperparameters and only disable the contribution under
test: \emph{single global prompt + multi-view} and \emph{single global
prompt + single view}. Decomposing the global prompt into three
\microtask{} prompts is the dominant gap: the single-prompt variants
fail to converge on the randomised protocol, whereas the per-\microtask{}
variant passes the rule-based PPO baseline.

\begin{figure}[!t]
\vspace{-2mm}
  \centering
  \includegraphics[width=0.9\linewidth]{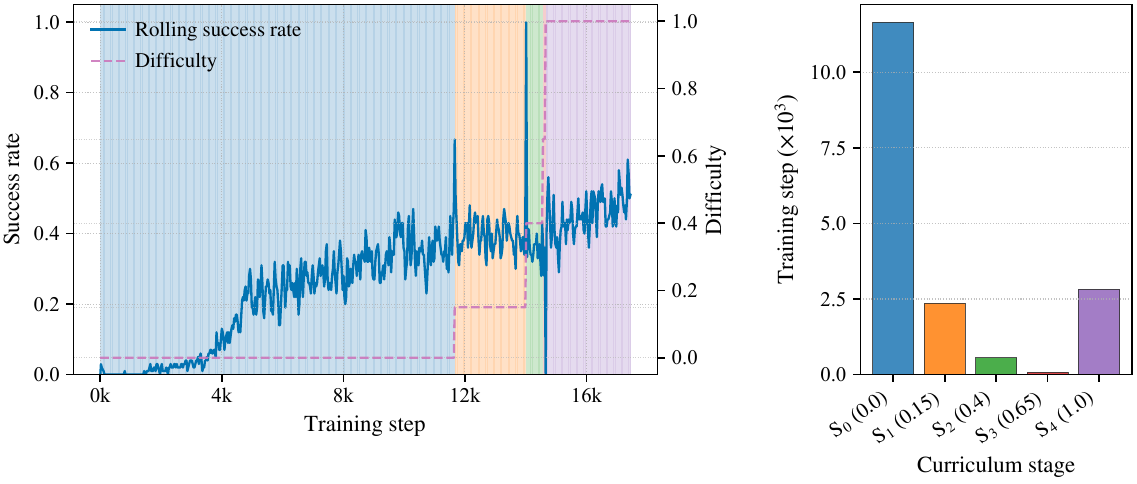}
  \caption{\textbf{Reverse-curriculum progression}: env steps per
    level and rolling per-level success rate during Stage-1 training.}
  \label{fig:rc_progression}
  \vspace{-4mm}
\end{figure}

\paragraph{Reverse-curriculum progression.}
\Cref{fig:rc_progression} shows env steps spent in each RC level and the
rolling per-level success rate during Stage-1 training. The pattern
that the agent dwells the longest at the hardest levels indicates the
randomised-init regime is the binding difficulty rather than any
intermediate one.

\begin{figure}[!b]
    \vspace{-4mm}
  \centering
  \includegraphics[width=0.9\linewidth]{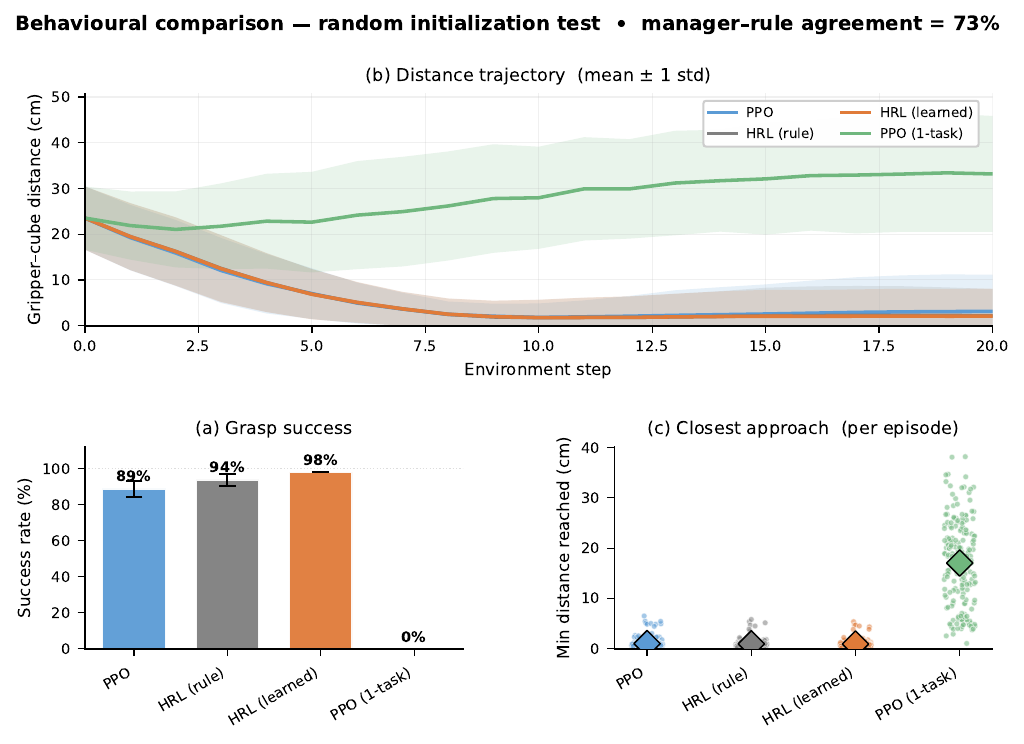}
  \caption{\textbf{Behavioural comparison on
    \texttt{FetchPickAndPlace-v4}} (1000 paired-seed episodes, random
    init). \emph{(a)} Grasp-detection success rate; \emph{(b--c)}
    behavioural breakdown over the same episodes. Manager--rule
    agreement: \textasciitilde$73\%$.}
  \label{fig:behavioural_4way}
\end{figure}

\paragraph{Behavioural comparison: rule vs.\ learned manager.}
\Cref{fig:behavioural_4way} compares four policies on Fetch under
random initial conditions. The two
hierarchical columns share the \emph{same} worker checkpoint and
differ only in the phase signal --- a distance-based rule vs.\ the
learned manager --- which isolates the effect of the manager from
any change in the worker. The full \rmtl{} stack with the learned
manager attains $\sim$$98\%$ success, up from $94\%$ for the
rule-based selector and $89\%$ for the non-hierarchical PPO worker,
while a single-prompt PPO baseline at the same compute fails
entirely ($0\%$). The learned manager agrees with the rule on $73\%$
of its decisions, yet still outperforms it: the rule is therefore
not optimal but a useful inductive bias, and using it as a BC
teacher (\Cref{app:impl}) lets the manager reach an end policy that
clears the rule's ceiling at a fraction of the training cost
required to learn phase selection from scratch.

\begin{figure}[!t]
  \centering
  \begin{subfigure}{0.49\linewidth}
    \centering
    \includegraphics[width=\linewidth]{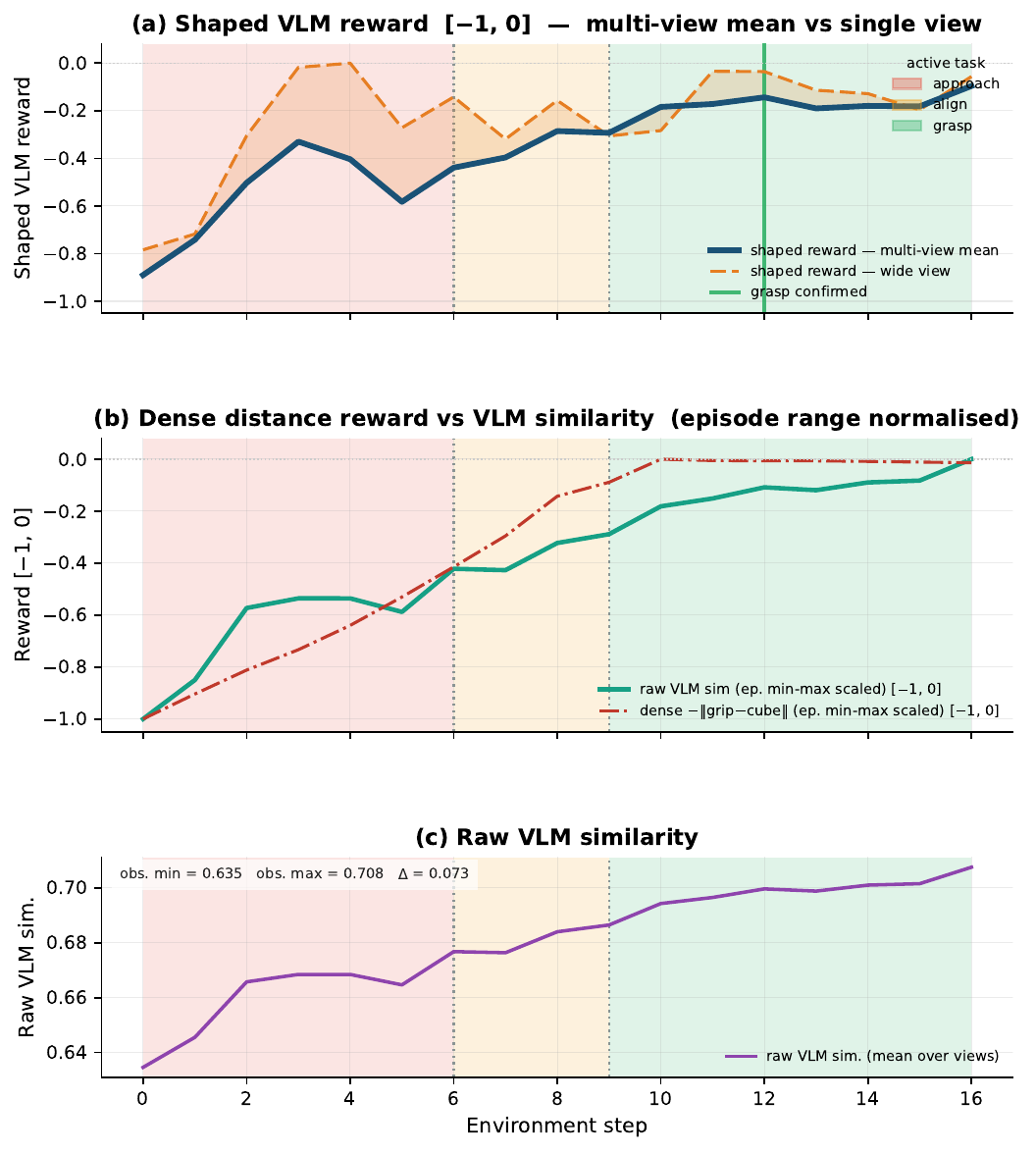}
    \caption{Episode 1.}
    \label{fig:multiview_story_ep01}
  \end{subfigure}\hfill
  \begin{subfigure}{0.49\linewidth}
    \centering
    \includegraphics[width=\linewidth]{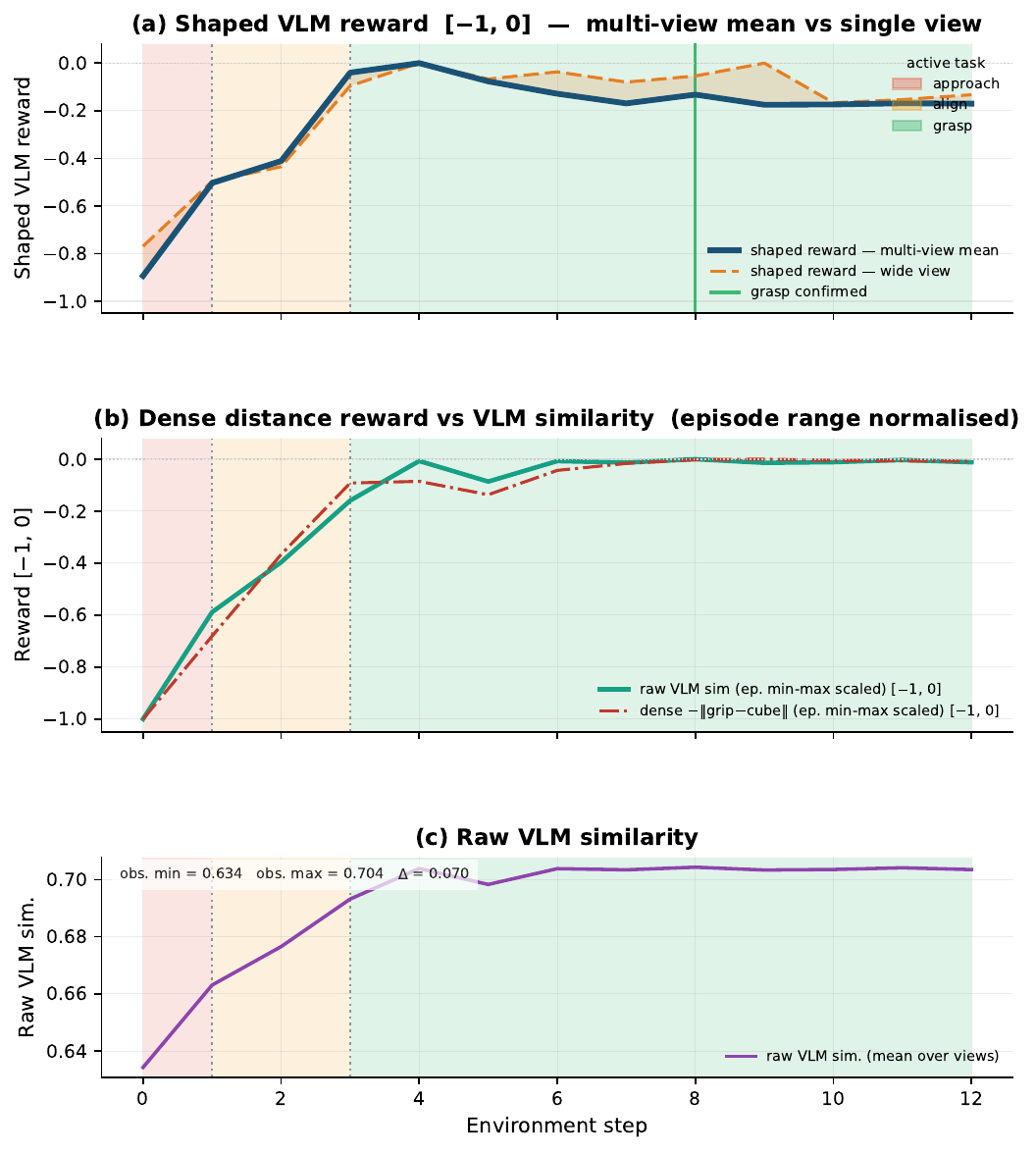}
    \caption{Episode 2.}
    \label{fig:multiview_story_ep02}
  \end{subfigure}
  \caption{\textbf{Multi-view reward over two episodes.} Each
    sub-figure shows, top to bottom: multi-view mean vs.\ single
    \emph{wide\_view} baseline; episode-rescaled VLM similarity vs.\ a
    dense $-\lVert g - o \rVert$ reward; raw VLM similarity.}
  \label{fig:multiview_story}
  \vspace{-5mm}
\end{figure}

\paragraph{Multi-view vs.\ single-view properties.}
\Cref{fig:multiview_story} traces the multi-view mean reward against a
single-view baseline (\emph{wide\_view}) over two independent eval
rollouts. In both episodes the multi-view mean is smoother than any
single view and preserves information about gripper configuration that
a hand-engineered dense distance reward does not.

\section{Discussion and limitations}
\label{sec:discussion}

\paragraph{Benefits of \rmtl{}.}
Three components account for most of the improvement over a
single-prompt VLM-RL baseline. \emph{(i)} The \microtask{}
decomposition converts a near-flat single-prompt VLM signal into a
piecewise-monotone signal that PPO can climb
(\Cref{fig:subtask_vs_onetask}). \emph{(ii)} Multi-view aggregation
reduces per-view occlusion noise and yields a smoother training
signal than any single view (\Cref{fig:multiview_story}).
\emph{(iii)} The reverse curriculum bridges the easy-init and
random-init regimes where the language reward alone is insufficient
to bootstrap exploration. The behavioural ablation in
\Cref{fig:behavioural_4way} further isolates the contribution of the
learned manager: replacing the
distance-based rule with a learned policy lifts random-init
grasp-detection success from $84\%$ to $\sim$$98\%$.

\paragraph{Limitations.}
The reward signal inherits the perceptual capabilities and
limitations of the underlying VLM: scenes whose lighting, shading,
or close-up texture fall outside the VLM's training distribution can
yield noisy similarity scores, and current VLMs exhibit limited
fine-grained spatial reasoning that can make per-step reward
ambiguous in contact-rich phases. The \emph{grasp} prompt also
naturally saturates once the cube is firmly held, reducing gradient
signal near the goal. We address the visual gap with a high-fidelity
rendering setup (\Cref{fig:domain_realization}) and the saturation
issue with light reward post-processing (\Cref{app:impl}); improving
robustness to varied visual conditions and stronger VLM spatial
reasoning remain open challenges.

\section{Future work}
\label{sec:future}

The most promising next step is to remove the last manual element in
\rmtl{} by pairing it with an \emph{LLM-based \microtask{}
disassembler}: given a high-level natural-language task description,
an LLM would automatically propose the stage-aligned prompt set and
a candidate phase-selection rule, replacing today's three
hand-written prompts and making the pipeline applicable to new tasks
without manual prompt engineering. We also plan to expand the
evaluation to additional manipulation tasks.

\bibliographystyle{plainnat}
\bibliography{references}

\newpage
\appendix

\section{Implementation details}
\label{app:impl}

\paragraph{Algorithm.} The RMTL algorithm is given below.
\begin{algorithm}[h!]
  \caption{Stage-2 HRL Learned Manager with a Frozen-then-Unfrozen PPO Worker}
  \label{alg:hrl-stage2}
  \begin{algorithmic}[1]
    \Require Pretrained PPO worker policy $\pi_\theta$, untrained manager policy $\pi_\psi$, rule classifier $\pi_{\text{rule}}$, VLM encoders $E,T$.
    \Require Hyperparameters: BC epochs $U_{BC}$, manager horizon $H$, exploration rate $\varepsilon$, freeze duration $T_{\text{freeze}}$, learning rates $\eta_{BC}, \eta_M, \eta_w'$.
    \Ensure Trained manager policy $\pi_\psi$ and fine-tuned worker policy $\pi_\theta$.
    
    \Statex \textbf{Offline Behavior Cloning (BC) Warmstart}
    \State Initialize BC dataset $\mathcal{D}_{BC} \gets \emptyset$
    \State Roll out $\pi_\theta$ guided by $\pi_{\text{rule}}$, record transitions $(s_t, \rho_t = \pi_{\text{rule}}(s_t))$ into $\mathcal{D}_{BC}$.
    \For{$u = 1, \dots, U_{BC}$}
      \State Sample batch $(s, \rho) \sim \mathcal{D}_{BC}$
      \State $\psi \gets \psi - \eta_{BC} \nabla_\psi \mathrm{CE}\!\bigl(\pi_\psi(\cdot \mid s),\, \rho\bigr)$ \Comment{Cross-entropy on rule labels}
    \EndFor
    
    \Statex \textbf{Online HRL Training}
    \State Freeze actor parameters $\theta_{\text{actor}}$ of $\pi_\theta$; keep critic parameters $\theta_{\text{critic}}$ trainable.
    \State Initialize rollout buffers $\mathcal{B}_{\text{worker}} \gets \emptyset$, $\mathcal{B}_{\text{manager}} \gets \emptyset$
    \For{env step $t = 0, 1, \dots$}
      \If{$t \bmod H = 0$ \textbf{or} \texttt{done}}
        \State Sample subtask $\phi_t \sim \varepsilon\text{-greedy}\bigl(\pi_\psi(\cdot \mid s_t),\, \varepsilon\bigr)$
      \Else
        \State $\phi_t \gets \phi_{t-1}$ \Comment{Hold subtask over horizon}
      \EndIf
      
      \State Construct augmented state $s_t' \gets (s_t,\, \mathrm{onehot}(\phi_t))$
      \State Sample primitive action $a_t \sim \pi_\theta(\cdot \mid s_t')$
      \State Step env, obtain multi-view shaped reward $\hat r_t$ via VLM encoders.
      \State Push $(s_t', a_t, \hat r_t)$ into $\mathcal{B}_{\text{worker}}$
      \State Accumulate segment reward: $R_{\text{seg}} \gets R_{\text{seg}} + \hat r_t$
      
      \If{manager segment is complete}
        \State Push $(s_{t-H}, \phi_t, \log \pi_\psi, \pi_{\text{rule}}(s_{t-H}), R_{\text{seg}})$ into $\mathcal{B}_{\text{manager}}$
        \State $R_{\text{seg}} \gets 0$
      \EndIf

      \If{$t = T_{\text{freeze}}$}
        \State Unfreeze actor parameters $\theta_{\text{actor}}$; set worker learning rate to $\eta_w'$.
      \EndIf
      
      \If{rollout complete}
        \State Update worker $\theta$ via PPO on $\mathcal{B}_{\text{worker}}$ \Comment{Updates critic only while $t < T_{\text{freeze}}$}
        
        \State Calculate mean $\mu_R$ and std $\sigma_R$ of segment returns in $\mathcal{B}_{\text{manager}}$
        \State Compute normalized advantages for each segment: $A_{\text{seg}} \gets \frac{R_{\text{seg}} - \mu_R}{\sigma_R}$
        
        \State $\mathcal{L}_M \gets -\,\mathbb{E}_{\mathcal{B}_{\text{manager}}}\!\left[ \log \pi_\psi(\phi \mid s) \cdot A_{\text{seg}} \right] - c_H\, \mathcal{H}[\pi_\psi] + w_{KL}\, \mathrm{KL}(\pi_{\text{rule}} \,\|\, \pi_\psi)$
        \State $\psi \gets \psi - \eta_M\, \nabla_\psi \mathcal{L}_M$
        
        \State Clear buffers $\mathcal{B}_{\text{worker}} \gets \emptyset$, $\mathcal{B}_{\text{manager}} \gets \emptyset$
      \EndIf
    \EndFor
  \end{algorithmic}
\end{algorithm}

\begin{figure}[h!]
  \centering
  \includegraphics[width=0.92\linewidth]{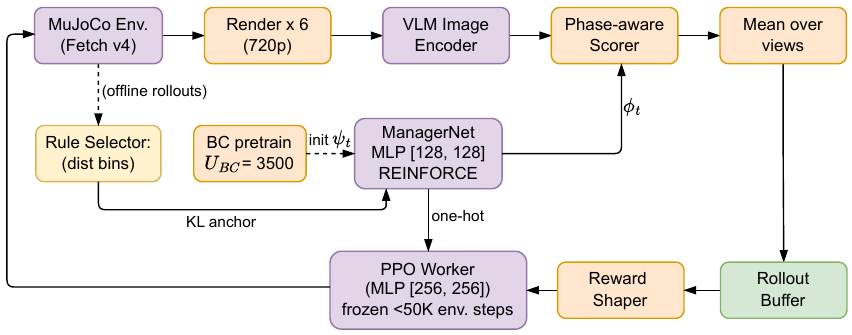}
  \caption{\textbf{Stage-2 HRL data flow with a frozen worker.} Solid
    arrows are online data; dashed arrows are offline / regularisation
    signals. ManagerNet (MLP $[128,128]$, REINFORCE-trained) is the
    only new learned component; the worker is loaded from the Stage-1
    PPO 560k checkpoint and its actor is frozen for the first
    $50{,}000$ env steps (critic remains trainable). The
    distance-based rule selector is used only offline (to label 250
    episodes for BC pretraining) and online as the KL anchor target
    in Eq.~\eqref{eq:manager-loss}; it never directly drives the
    prompt during Stage-2 training.}
  \label{fig:hrl_flow}
\end{figure}

\paragraph{Stage-2 data flow.}
\Cref{fig:hrl_flow} sketches the full Stage-2 wiring. ManagerNet drives
both the worker (via a phase-onehot input) and the phase-aware scorer
(via the active prompt index), while the distance-based rule classifier
is retained \emph{only} as an offline BC label source and as the
KL-anchor target during REINFORCE.

\paragraph{BC warmstart.} Before any RL update, we run a
behaviour-cloning phase: the rule-based phase classifier is rolled out
for 250 episodes on the loaded PPO 560k checkpoint, producing
$(s, i_{\text{rule}})$ pairs. The manager is then fit to these labels
with cross-entropy for 3500 updates at batch size 64.

\paragraph{REINFORCE on segment returns.} After warmstart, the manager
is trained at the end of every PPO rollout with a standard REINFORCE
objective on segments of length $H$:
\begin{align}
  \mathcal{L}_M
  &= -\,
       \mathbb{E}_{\text{seg}}\!\left[
         \log \pi_M(\phi \mid s_{\text{seg}}) \cdot A_{\text{seg}}
       \right]
       \;-\; c_H \mathcal{H}\!\bigl[\pi_M\bigr]
       \;+\; w_{\text{KL}} \,\mathrm{KL}\!\bigl(\pi_{\text{rule}} \,\|\, \pi_M\bigr),
  \label{eq:manager-loss}
\end{align}
where $A_{\text{seg}}$ is the rollout-internal $z$-score of the
segment return $\sum_{t \in \text{seg}} r_t$, $\mathcal{H}[\cdot]$ is
the distributional entropy with bonus $c_H = 0.01$, and the third
term is a KL anchor to the rule-based policy with weight
$w_{\text{KL}} = 0.30$. Action selection uses $\varepsilon$-greedy
exploration with $\varepsilon = 0.20$. The optimizer is Adam at
learning rate $3 \times 10^{-4}$.

\paragraph{Frozen worker.} Throughout this paper, Stage 2 starts from
a specific Stage-1 checkpoint (PPO 560k env steps) loaded into the
worker. For the first $50{,}000$ environment steps the worker actor's
parameters are frozen
($\texttt{requires\_grad} = \mathrm{False}$ on features
+ \texttt{mlp\_extractor.policy\_net}
+ \texttt{action\_net} + \texttt{log\_std}), keeping only the critic
trainable. After 50k steps a callback unfreezes the actor and reduces
the worker learning rate to $1 \times 10^{-4}$. The BC labels above
are also generated against this same 560k checkpoint for
distributional consistency. Freezing the worker stabilises the
manager's REINFORCE target during the most non-stationary phase of
training, which we found important for avoiding mode collapse on the
manager's softmax.

\paragraph{Phase prompts.}
The three Stage-1 \microtask{} target prompts and the shared baseline
prompt are written once and reused across runs. They are:
\begin{itemize}\itemsep1pt
  \item baseline $b$: \emph{``a robot gripper and a red cube on a table''};
  \item \emph{approach} target $\tau_{\text{approach}}$: \emph{``a
    two-finger robot gripper with open jaws centered above a red cube''};
  \item \emph{align} target $\tau_{\text{align}}$: \emph{``a robot
    gripper pinching a red cube between two fingers''};
  \item \emph{grasp} target $\tau_{\text{grasp}}$: \emph{``a robot
    gripper lifting a red cube between its fingers''}.
\end{itemize}
The rule-based selector (Stage~1) chooses the active \microtask{}
deterministically from a simple geometric feature; the learned manager
(Stage~2) replaces this rule with a small policy over the same three
prompts.

\paragraph{Camera bank.}
The six camera viewpoints used by all multi-view configurations are
defined as MuJoCo free-camera settings (azimuth, elevation, distance,
lookat). Their parameters are:
\emph{wide\_view} $(132, -14, 1.5)$, \emph{front} $(180, -20, 1.1)$,
\emph{front\_side} $(45, -30, 1.2)$, \emph{isometric} $(225, -35, 1.1)$,
\emph{side} $(90, -15, 1.2)$, \emph{close\_up} $(135, -25, 0.8)$, all
looking at approximately $(1.3, 0.75, 0.42)$. Realistic-visuals shading
is on.

\paragraph{Termination bonus.}
An episode terminates with success once the gripper is within $0.025$\,m
of the cube and the cube has been lifted at least $0.02$\,m above the
table for two consecutive steps; on success we add a one-shot reward of
$+1.0$ to the final step.

\paragraph{Reverse-curriculum thresholds.}
The 5 RC stages have $(d, [d_{\min}, d_{\max}], \theta_{\text{succ}},
n_{\min})$ values
$(0,\,[0,0.08],\,0.5,\,50)$,
$(0.15,\,[0.08, 0.25],\,0.5,\,80)$,
$(0.4,\,[0.25, 0.50],\,0.45,\,120)$,
$(0.65,\,[0.50, 0.75],\,0.4,\,150)$, and
$(1.0,\,[0.75, 1.0],\,-,\,0)$ (the final stage is ungated).

\paragraph{Manager hyperparameters.}
Manager horizon $H = 3$, hidden sizes $[128, 128]$, learning rate
$3 \times 10^{-4}$, entropy coefficient $0.01$, KL-to-rule weight
$0.30$, $\varepsilon$-greedy $0.20$, BC pretrain updates $3500$, BC
batch size $64$, BC pretrain episodes $250$, BC pretrain checkpoint
PPO 560k, frozen-worker until 50k env steps, unfreeze learning rate
$1 \times 10^{-4}$.

\section{Hyperparameters}
\label{app:hyperparameters}
\paragraph{Stage 1 hyperparameters.} Stage 1 training hyperparameters are given in the table below.
\begin{table}[h!]
  \centering
  \caption{Stage-1 training setup: environment, VLM scorer, phase
    prompts, reward shaping, reverse curriculum, and PPO worker
    hyperparameters. Identical across all configs in
    \Cref{sec:expt-results} except where the ablation explicitly
    disables a feature.}
  \label{tab:hparams_stage1}
  \small
  \begin{tabular}{l l l}
    \toprule
    \textbf{Component} & \textbf{Hyperparameter} & \textbf{Value} \\
    \midrule
    \multirow{8}{*}{Environment}
      & task / version              & \texttt{FetchPickAndPlace-v4} (Gymnasium-Robotics) \\
      & physics                     & MuJoCo \\
      & state observation           & 25-D (gripper pose, cube pose, fingers, goal) \\
      & action space                & 4-D continuous (gripper $\Delta xyz$ + finger) \\
      & episode length              & 40 steps \\
      & parallel envs per learner   & 16 \\
      & workspace box $(x, y, z)$   & $[1.00, 1.60] \times [0.35, 1.15] \times [0.43, 0.90]$\,m \\
      & success criterion           & $\lVert g - o \rVert \le 0.025$\,m and lift $\ge 0.02$\,m for 2 steps; $+1.0$ bonus \\
    \midrule
    \multirow{5}{*}{VLM scorer}
      & backbone                    & PE-Core-bigG-14-448 \citep{tschannen2025pecore} \\
      & view aggregation            & mean over $V {=} 6$ cameras \\
      & render resolution           & $720 \times 720$ \\
      & projection mix $\alpha$     & 0.9 \\
      & VLM batch size              & 64 \\
    \midrule
    \multirow{2}{*}{Phase prompts}
      & \# \microtasks{} $K$        & 3 (\emph{grasp} / \emph{align} / \emph{approach}) \\
      & distance bins               & $(0,\,0.04],\ (0.04,\,0.15],\ (0.15,\,\infty)$\,m \\
    \midrule
    \multirow{5}{*}{Reward shaping}
      & initial min / max           & 0.45 / 0.70 \\
      & adaptation rate / window    & 0.01 / 3 steps \\
      & trend / momentum weights    & 0.3 / 0.3 \\
      & drop-penalty multiplier     & $\times 2$ \\
      & output range                & $[-1, 0]$ \\
    \midrule
    \multirow{4}{*}{Reverse curriculum}
      & mode / window               & scalar / 100 episodes \\
      & stages $d$                  & $\{0,\ 0.15,\ 0.4,\ 0.65,\ 1.0\}$ \\
      & success thresholds          & $\{0.5,\ 0.5,\ 0.45,\ 0.4,\ -\}$ \\
      & min episodes per stage      & $\{50,\ 80,\ 120,\ 150,\ 0\}$ \\
    \midrule
    \multirow{10}{*}{PPO worker}
      & policy / network            & MultiInputPolicy, MLP $[256, 256]$ for $\pi$ and $V$ \\
      & rollout length $n_{\text{steps}}$, $\texttt{train\_freq}$ & 40 \\
      & batch size                  & 64 \\
      & epochs per update           & 10 \\
      & learning rate               & $3 \times 10^{-4}$ \\
      & discount $\gamma$           & 0.95 \\
      & GAE $\lambda$               & 0.95 \\
      & clip range                  & 0.2 \\
      & entropy coef                & 0.01 \\
      & vf coef / max grad norm     & 0.5 / 0.5 \\
    \bottomrule
  \end{tabular}
\end{table}

\paragraph{Stage 2 hyperparameters.} Stage 2 training hyperparameters are given in the table below.

\begin{table}[h!]
  \centering
  \caption{Stage-2 manager hyperparameters for the frozen-worker setup.}
  \label{tab:hparams_stage2}
  \small
  \begin{tabular}{l l l}
    \toprule
    \textbf{Component} & \textbf{Hyperparameter} & \textbf{Value} \\
    \midrule
    \multirow{8}{*}{Manager $\pi_\psi$}
      & architecture           & MLP $[128, 128]$ + softmax \\
      & input                  & flattened state (25-D) \\
      & output                 & $K$-way categorical ($K{=}3$) \\
      & shared across envs     & yes \\
      & manager horizon $H$    & 3 env steps \\
      & optimizer              & Adam, lr $\eta_M = 3 \times 10^{-4}$ \\
      & exploration            & $\varepsilon$-greedy with $\varepsilon = 0.20$ \\
      & manager drives prompt  & yes \\
    \midrule
    \multirow{3}{*}{REINFORCE objective}
      & advantage              & rollout-internal z-score of segment returns \\
      & entropy coef $c_H$     & 0.01 \\
      & KL-to-rule weight $w_{KL}$ & 0.30 \\
    \midrule
    \multirow{5}{*}{BC warmstart (offline)}
      & teacher                & rule-based phase classifier $\pi_{\text{rule}}$ (distance bins) \\
      & source ckpt for rollouts & PPO 560k env steps \\
      & rollout episodes       & 250 \\
      & CE updates             & 3500, batch size 64 \\
      & pre-fill segments      & 512 (before first BC step) \\
    \midrule
    \multirow{5}{*}{Frozen-worker gate}
      & worker init ckpt       & PPO 560k env steps (same as BC source) \\
      & freeze duration $T_{\text{freeze}}$ & 50{,}000 env steps \\
      & frozen tensors         & features, \texttt{policy\_net}, \texttt{action\_net}, $\log\sigma$ \\
      & critic during freeze   & trainable \\
      & post-unfreeze worker lr $\eta_w'$ & $1 \times 10^{-4}$ \\
    \bottomrule
  \end{tabular}
\end{table}

\section{Projection-mix sweep}
\label{app:alpha-sweep}

This appendix expands on the choice of the projection-mix scalar
$\alpha = 0.9$ used in Eq.~\eqref{eq:projection}.

\begin{figure}[h!]
  \centering
  \includegraphics[width=0.95\linewidth]{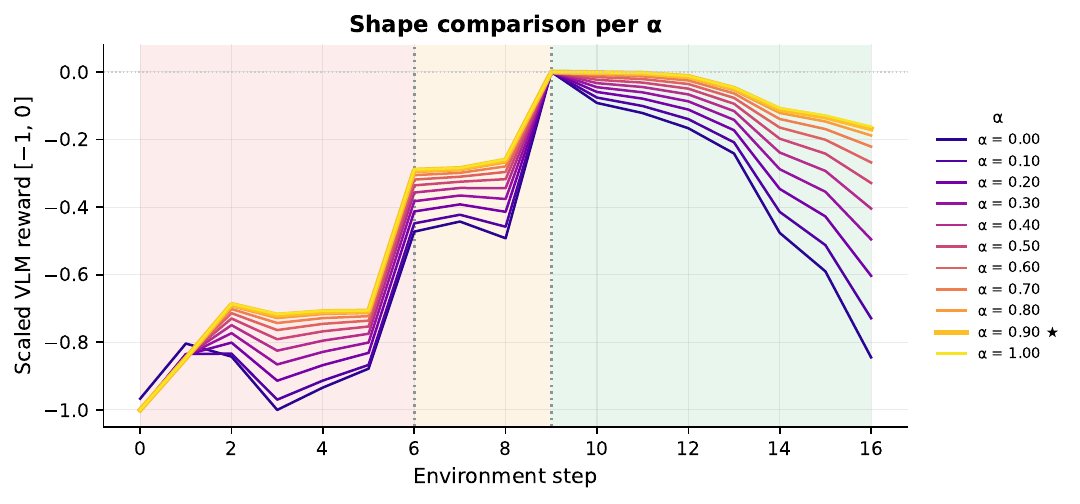}
  \caption{\textbf{Effect of $\alpha$ on the temporal reward profile,
    one representative episode.} (a) Raw active-\microtask{} VLM
    reward for $\alpha \in [0, 1]$: higher $\alpha$ stretches the
    reward to a larger absolute range (better signal-to-noise) but
    does not change the \emph{shape} of the trajectory. (b) Each
    curve rescaled to $[-1, 0]$ via per-$\alpha$ min--max: the
    rescaled trajectories cluster tightly, confirming that $\alpha$
    primarily controls magnitude rather than the temporal structure
    of the gradient. Background bands mark active \microtasks{};
    $\alpha = 0.9$ (highlighted with $\star$) is our default.}
  \label{fig:alpha_timeseries}
\end{figure}

\paragraph{Interpreting $\alpha$.}
The scalar $\alpha \in [0,1]$ interpolates between unprojected cosine
similarity ($\alpha{=}0$) and projection strictly along the
target--baseline axis ($\alpha{=}1$). As $\alpha$ increases, the raw
per-step VLM reward typically spans a wider numerical range for the
\emph{same} underlying trajectory: the temporal profile---which
\microtask{} is active and how similarity evolves---remains largely
unchanged, but PPO sees larger step-to-step differences and therefore
a stronger gradient signal. \Cref{fig:alpha_timeseries} summarises this
on one episode: all $\alpha$ curves follow the same phase structure,
but higher $\alpha$ stretches the vertical extent.

\begin{figure}[h!]
  \centering
  \includegraphics[width=0.9\linewidth]{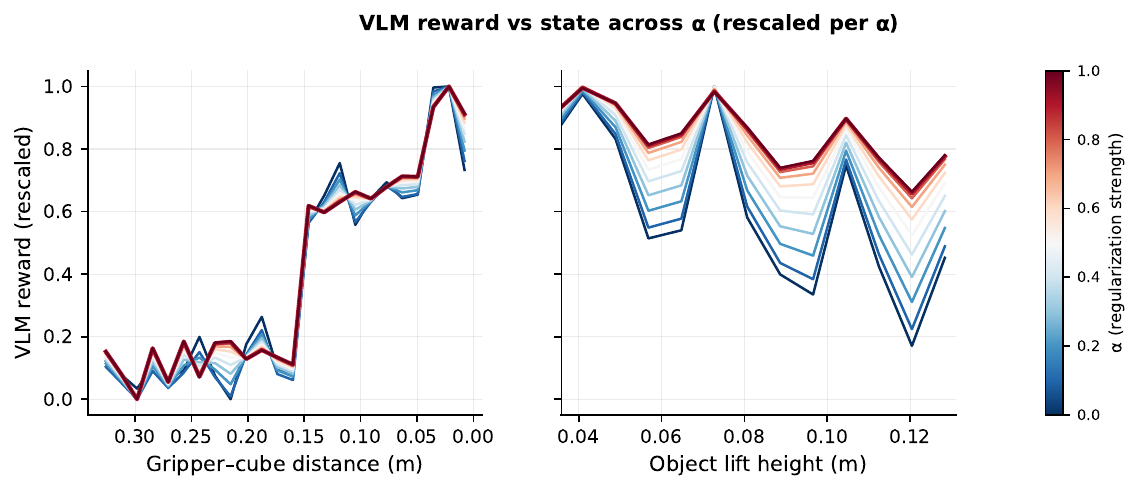}
  \caption{\textbf{Effect of $\alpha$ on the state-conditioned reward,
    aggregated over 10 analysis episodes.} Each curve is the per-bin
    mean of the rescaled active-\microtask{} VLM reward; one curve
    per $\alpha \in \{0, 0.1, \ldots, 1\}$ (colourbar). \emph{Left:}
    monotone climb in gripper--cube distance --- the dominant state
    dimension --- with the $\alpha$ family tightly bundled.
    \emph{Right:} object-lift height. The sharp drop at $h \approx 0$
    is the table--air boundary; the oscillatory plateau for $h > 0$
    sits at the saturation level of the \emph{grasp} prompt, which is
    the terminal \microtask{} in our ordering and therefore stays
    active for all $h > 0$. Variation along the plateau is
    view-dependent VLM noise within a single saturated prompt rather
    than a learnable gradient in the lift dimension.}
  \label{fig:alpha_sweep_state}
\end{figure}

\Cref{fig:alpha_sweep_state} aggregates the same episodes over two
scalar state coordinates. In gripper--cube distance (left), reward
climbs monotonically as the gripper approaches the cube; the family of
$\alpha$ curves overlaps, so $\alpha$ does not flip \emph{which}
physical variable the reward tracks. In object lift height (right),
the on-table ($h{\approx}0$) vs.\ lifted ($h{>}0$) transition marks
the switch into the terminal \emph{grasp} \microtask{}; after that
the prompt is saturated, so variation along height is dominated by
visual noise rather than an informative gradient. Together, these
plots justify $\alpha{=}0.9$ as a high-amplitude setting that
preserves the phase structure, and they motivate keying the
rule-based selector on distance rather than lift height
(\S\ref{sec:rule-worker}).

\end{document}